\pdfoutput=1  

\documentclass[letterpaper, 10 pt, conference]{ieeeconf}  

\IEEEoverridecommandlockouts                              

\usepackage{graphicx} 
\usepackage{amsmath} 
\usepackage{amssymb}  
\usepackage{float} 
\usepackage[section]{placeins} 
\usepackage{cite}
\usepackage[hidelinks]{hyperref}
\usepackage{algorithm}
\usepackage{algorithmic}
\usepackage{multirow}
\let\labelindent\relax
\usepackage{enumitem}
\usepackage{needspace}
\usepackage{booktabs}
\usepackage{makecell}
\usepackage{caption} 

\newcommand{\best}[1]{\textbf{#1}}
\newcommand{\second}[1]{\underline{#1}}

\usepackage{titlesec}
\titlespacing*{\section}{0pt}{6pt plus 2pt minus 2pt}{3pt plus 1pt minus 1pt}
\titlespacing*{\subsection}{0pt}{4pt plus 2pt minus 1pt}{2pt plus 1pt minus 1pt}
\AtBeginDocument{%
  \abovedisplayskip=4pt plus 1pt minus 1pt
  \belowdisplayskip=4pt plus 1pt minus 1pt
  \abovedisplayshortskip=0pt plus 1pt
  \belowdisplayshortskip=0pt plus 1pt
}

\title{\LARGE \bf
IMLE-VLA: Fast Single-Step Action Generation for Vision-Language-Action Policies
}

\author{Kian Hosseinkhani$^{1*}$ \quad Qinhe Peng$^{2}$ \quad George Shramko$^{1}$ \quad Mehran Aghabozorgi$^{1}$ \quad Jianing Qian$^{2}$\\
Tristan Engst$^{1}$ \quad Alireza Moazeni$^{1}$ \quad Dinesh Jayaraman$^{2}$ \quad Ke Li$^{1,3,4}$\\[4pt]
{\normalsize $^{1}$Simon Fraser University \quad $^{2}$University of Pennsylvania}\\
{\normalsize $^{3}$Alberta Machine Intelligence Institute (Amii) \quad $^{4}$Canada CIFAR AI Chair}\\
{\small $^{*}$Corresponding author}}

\makeatletter
\def\bstctlcite#1{\@bsphack
  \@for\@citeb:=#1\do{%
    \edef\@citeb{\expandafter\@firstofone\@citeb}%
    \if@filesw\immediate\write\@auxout{\string\citation{\@citeb}}\fi}%
  \@esphack}
\makeatother

\begin{document}
\bstctlcite{IEEEexample:BSTcontrol}

\twocolumn[{%
\renewcommand\twocolumn[1][]{#1}%
\maketitle
\thispagestyle{empty}
\begin{center}
\centering
\includegraphics[width=1\textwidth]{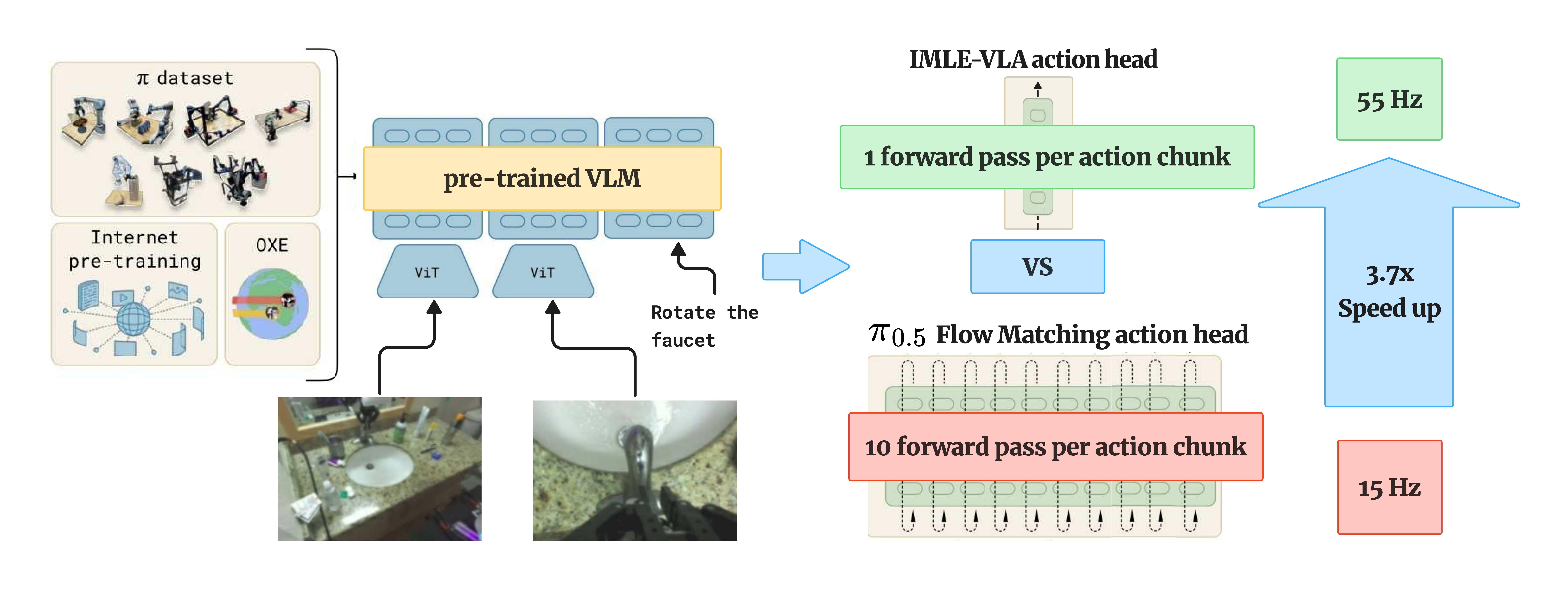}
{\captionsetup{hypcap=false}%
\captionof{figure}{Overview of IMLE-VLA: drop in single-step action head removing iterative modelling in VLAs}%
\label{fig:main}}
\end{center}
\vspace{0em}
}]
\pagestyle{empty}

\begin{abstract}
Vision-language-action (VLA) policies leverage pretrained vision-language backbones to achieve strong cross-task generalization. A leading design couples this backbone with a dedicated continuous action head trained via diffusion or flow matching. However, such heads rely on iterative multi-step sampling, for example 10 Euler steps in $\pi_{0.5}$. This creates an inference bottleneck that produces stop-and-go movement in the robot and slower task completion. We introduce \textbf{IMLE-VLA}, which replaces the iterative action head with a single-step conditional generator trained via conditional Implicit Maximum Likelihood Estimation (cIMLE). The cIMLE objective promotes multimodal action coverage, avoiding the mode collapse of naive regression heads while eliminating multi-step sampling entirely. When IMLE-VLA is applied to $\pi_{0.5}$, it increases inference frequency $3.67\times$ ($55$\,Hz vs.\ $15$\,Hz), enabling up to $11\times$ higher action throughput. On the 40-task LIBERO benchmark, IMLE-VLA achieves the highest average success rate ($98.0\%$) among all baselines while leading in inference frequency. Under the test-time perturbations of LIBERO-plus, IMLE-VLA retains $\pi_{0.5}$'s robustness while other baselines degrade sharply, confirming that the cIMLE head preserves generalization. Real-world experiments on a Franka Emika Panda across four tasks demonstrate smoother motion ($2.2\times$--$3.0\times$ lower jerk) and faster task completion, with IMLE-VLA outperforming $\pi_{0.5}$ on every task and reducing average VLA inference time per episode by $3.9\times$--$6.6\times$. Videos and code are available at \url{https://kianhk6.github.io/IMLE-VLA/}.
\end{abstract}

\section{Introduction}\label{sec:introduction}

Vision-language-action (VLA) policies build on pretrained vision-language models (VLMs), inheriting internet-scale visual and linguistic knowledge that enables strong generalization across tasks and embodiments~\cite{reed2022generalist,openxembodiment2023,brohan2023rt2,kim2024openvla}, making them the dominant generalist paradigm in robot learning. However, this generalization comes at a steep computational cost. Large vision-language backbones combined with iterative action-sampling procedures make VLA inference a critical bottleneck for high-frequency real-time robot control~\cite{vlaperf2025,pertsch2025fast,jeon2025shallow,kim2025openvlaoft,zhang2025quantvla,chen2025rlrc,hirose2025asyncvla,ma2025realtime}. The resulting latency introduces dead time in the control loop: under synchronous execution, the robot stalls until the next prediction is ready, producing stop-and-go execution and longer task completion times.

This bottleneck primarily stems from the way VLAs generate actions, which follows one of two dominant paradigms: autoregressive decoding of discrete tokens and multi-step sampling of continuous actions. Autoregressive VLAs represent each action as a sequence of discrete tokens and decode them one at a time~\cite{openxembodiment2023,brohan2023rt2,kim2024openvla}. Furthermore, their reliance on discrete tokenization can limit the fine-grained continuous precision needed for dexterous manipulation~\cite{black2024pi0,black2025pi05}. The second class generates continuous actions through multi-step sampling with diffusion or flow matching, typically via an action expert attached to a VLM backbone such as $\pi_{0.5}$~\cite{black2025pi05}. Here the model predicts a chunk of future actions together, yet producing each chunk still requires multiple sequential forward passes (e.g., $10$ Euler steps)~\cite{black2024pi0,black2025pi05,ghosh2024octo,liu2024rdt1b,wang2024cogact,wang2024hpt}. Consequently, action generation remains a primary bottleneck for real-time control.

One of the key motivations for diffusion and flow-matching action heads is their capacity to handle the inherent multimodality of robot action distributions~\cite{black2024pi0,black2025pi05}. A single instruction such as ``clear the table'' can be executed through many valid action sequences, and these models aim to capture this complexity. A naive solution would be to eliminate sequential generation via a single-step regression head. However, such models are prone to \textit{mode collapse}~\cite{chi2023diffusionpolicy,shafiullah2022bet,black2024pi0}: the regression objective averages over valid modes, pulling the prediction toward a single mean behavior that degrades task performance when the action distribution is multimodal. We empirically validate this effect in Sec.~\ref{sec:method_analysis}.

This tension between expressivity and efficiency raises a central question: \textit{can a single-step generator with a mode-covering objective replace iterative action generation while improving task performance?}

In this work, we introduce \textbf{IMLE-VLA}, a single-step action generation framework for VLAs that replaces multi-step sampling with a conditional Implicit Maximum Likelihood Estimation (cIMLE)~\cite{imle,aghabozorgi2023adaptive,peng2022chimle,vashist2024rsimle,aghabozorgi2026wimle} objective that explicitly optimizes for mode coverage.

We instantiate IMLE-VLA by replacing the 10-step flow-matching action head of $\pi_{0.5}$~\cite{black2025pi05} with our single-step generator. As the modification is confined to the action head, the full VLM backbone is preserved unchanged. This yields a $3.67\times$ increase in inference frequency ($55$\,Hz vs.\ $15$\,Hz). Combined with an extended execution horizon $H$, which lets the policy replan less frequently, this compounds into up to $11.0\times$ higher action throughput (inference frequency $\times$ $H$)~\cite{kim2025openvlaoft,pertsch2025fast,jeon2025shallow,ma2025realtime,hirose2025asyncvla}. This directly targets the ``stop-and-go'' control paradigm~\cite{brohan2023rt2,kim2024openvla,zhao2023act,black2024pi0,black2025pi05,ghosh2024octo,pertsch2025fast} that high VLA latency imposes: the robot idles while the model computes, producing visible pauses and stuttering motion. By cutting the time the robot waits on inference, IMLE-VLA reduces these pauses and reacts faster in dynamic settings, leading to faster task completion.

We evaluate IMLE-VLA in both simulation and the real world. On the \textbf{LIBERO simulation benchmark}~\cite{liu2023libero} (40 tabletop manipulation tasks across four suites), IMLE-VLA achieves the highest average success rate ($98.0\%$) among all compared methods. To probe whether these gains hold under distribution shift, we further evaluate IMLE-VLA on \textbf{LIBERO-plus}~\cite{fei25libero-plus}, which applies systematic test-time perturbations to the LIBERO tasks. Under these conditions, IMLE-VLA retains $\pi_{0.5}$'s robustness while all other baselines degrade sharply. In a \textbf{real-world setting} with a Franka Emika Panda, we test four tasks spanning single-step manipulation, multi-stage sequential reasoning, and reactivity to a dynamic environment. IMLE-VLA outperforms $\pi_{0.5}$ on every task while producing measurably smoother motion, reducing proprioceptive jerk by $2.2\times$--$3.0\times$. Across both settings, we measure the wall-clock time spent on VLA forward passes per episode, excluding robot execution and all other non-inference overhead. IMLE-VLA reduces this by up to $10.5\times$ in simulation and up to $6.6\times$ on the real robot. 

\section{Related work}\label{sec:related}
Our work targets the sequential action-generation bottleneck in generalist VLAs. We first review complementary system-level optimization strategies, and then review distillation and teacher-free single-step methods.

\textbf{System-level optimizations.}
Recent work~\cite{ma2025realtime} uses hardware-specific kernel tuning to accelerate inference, yet latency remains bounded by the sequential steps of diffusion-based heads. IMLE-VLA eliminates this bottleneck via single-step generation. These system-level gains are complementary to our algorithmic approach; by leveraging their efficient Triton kernels, we further amplify our inference frequency as demonstrated in Sec.~\ref{sec:experiments}.

\textbf{Distillation methods.}
One line of work accelerates VLAs by training a faster \emph{student} to imitate a pretrained \emph{teacher}, which ties the student's behavior to the teacher's output distribution. Shallow-$\pi$~\cite{jeon2025shallow} distills $\pi_{0.5}$ into a shallower, lower-capacity student, and the resulting speedup comes at a measurable cost in success rate relative to the teacher (Sec.~\ref{sec:experiments}). A separate thread distills a single-step student from a multi-step teacher~\cite{prasad2024consistency,wang2024onedp,Dong2026FromFT}, leaving the student with the biases of the teacher's iterative sampler. Unlike distillation-based accelerators, IMLE-VLA's action head has no teacher: it is fit directly to expert actions under a cIMLE objective. Its outputs are therefore not capped by a teacher's performance, nor shaped by an iterative sampler's biases.

\textbf{Teacher-free single-step methods.}
A separate line of work sidesteps the teacher entirely, learning single-step action generation directly from expert data. Among VLAs, the prominent example is OpenVLA-OFT~\cite{kim2025openvlaoft}, which replaces the autoregressive decoding of OpenVLA with an L1 regression head. L1 regression is known to collapse to the conditional median under multimodal action distributions~\cite{chi2023diffusionpolicy,shafiullah2022bet,black2024pi0}, and OpenVLA-OFT attributes L1's viability to the capacity of OpenVLA's 7B backbone~\cite{kim2025openvlaoft}. This reliance on a 7B backbone leaves OpenVLA-OFT $5.3\times$ slower than IMLE-VLA despite IMLE-VLA's smaller 3B backbone (Table~\ref{tab:libero_success_horizon}). Yet this backbone scale does not confer robustness under distribution shift: on LIBERO-plus, OpenVLA-OFT's success rate degrades sharply under test-time perturbations (Fig.~\ref{fig:libero_plus}). IMLE-VLA, by contrast, remains robust under these perturbations, despite its smaller 3B backbone.

\textbf{From specialists to a generalist.}
Across both lines of work above, prior efforts at single-step multimodal action generation have largely been confined to specialist policies~\cite{prasad2024consistency,wang2024onedp,frans2025shortcut}, task-specific models that require expensive retraining and data collection for every new setting. Even in these narrow domains, mitigating mode collapse requires complex auxiliary regularizers that encourage distribution coverage~\cite{zou2025dm1,dmpo2025}. The cIMLE objective~\cite{imle} instead enforces mode coverage \emph{by construction}, but its application in robotics has been limited to specialist regimes~\cite{rana2025imlepolicy,Bhaskar2026PRISMPR,Dong2026FromFT,Lee2026ImplicitML} and thus inherits the same generalization bottleneck. IMLE-VLA is the first to adapt cIMLE to VLAs, achieving single-step action generation in a generalist policy that leverages internet-scale vision-language pretraining to generalize across tasks and embodiments.

\section{Method}\label{sec:method}
We present \textbf{IMLE-VLA}, a framework for single-step, multimodal action generation in vision-language-action (VLA) policies. Our approach eliminates the computational bottleneck of multi-step iterative steps by training the action head with a conditional Implicit Maximum Likelihood Estimation (cIMLE) objective. Below, we detail our architectural framework and the mathematical formulation of cIMLE. Specifically, we first describe the coupling of a frozen VLM backbone with a lightweight, single-step action generator (\ref{sec:arch}); then, we formalize the cIMLE training objective and its unique properties for modeling multimodal robotic behaviors (\ref{sec:cimle}).

\subsection{Single-Step VLA Architecture}\label{sec:arch}
We operate within the VLA paradigm by coupling a high-capacity, frozen VLM backbone with a lightweight action head~\cite{black2024pi0,black2025pi05,ghosh2024octo,wang2024cogact}. By fixing the VLM parameters, we retain the robust spatial and semantic features acquired during internet-scale pretraining. Since the VLM constitutes the vast majority of model parameters and remains frozen, training reduces to the action head alone, keeping the cost minimal. 

At each timestep $t$, the observation $o_t$ consists of the visual inputs and language instruction. The VLM encodes it into a set of latent embeddings $f_{\mathrm{VLM}}(o_t)$, which condition the action generation process. Unlike standard iterative methods such as diffusion or flow-matching, which require multiple sequential denoising steps, our approach uses a single-step conditional generator $G_\theta$. This generator directly maps Gaussian noise to multimodal action chunks, effectively mitigating a significant latency contributor in iterative VLA inference.

Formally, we represent the generator output as an action chunk $A \in \mathbb{R}^{C \times D}$, where $C$ denotes the prediction horizon and $D$ is the action-space dimensionality. 
To generate a candidate, we sample a latent noise vector $z \sim \mathcal{N}(0, \mathbf{I})$ of the same dimensionality as $A$. The generator $G_\theta$ then maps the VLM embedding $f_{\mathrm{VLM}}(o_t)$ and noise $z$ to a complete action chunk in a single forward pass:
\begin{equation}
\hat{A} = G_{\theta}(f_{\mathrm{VLM}}(o_t), z).
\end{equation}
By sampling different latent vectors $z$, $G_\theta$ can produce diverse action chunk candidates that capture the underlying conditional distribution $p(A \mid o_t)$. This stochasticity ensures the model avoids regressing to a single average outcome, preserving the expressivity required for complex behaviors.
We initialize $G_{\theta}$ from a pretrained $\pi_{0.5}$ checkpoint and fine-tune the action head using the cIMLE objective (\ref{sec:cimle}) while the VLM backbone remains frozen.

\subsection{Conditional IMLE (cIMLE)}\label{sec:cimle}
The pretrained $\pi_{0.5}$ action head is trained via Conditional Flow Matching (CFM)~\cite{lipman2023flow,black2024pi0}, which necessitates the numerical integration of a learned velocity field during inference. In practice, this requires multiple sequential forward passes, typically 10 Euler steps, to generate a single action chunk, introducing a considerable inference-time integration bottleneck.

To eliminate this bottleneck, we seek to generate each action chunk in a single forward pass without sacrificing the multimodality of the action distribution. As discussed in Sec.~\ref{sec:introduction}, a single instruction can be executed through many valid action sequences, and each of these behaviors is a distinct mode of the conditional action distribution $p(A \mid o_t)$. Although each training pair $(o_t^{(i)}, A_{\mathrm{gt}}^{(i)})$ captures only one of these behaviors, the data contains observations that are very similar yet correspond to very different ground-truth actions. To preserve this multimodality, a faster action head must therefore remain capable of assigning different actions to the same observation.

The conditional IMLE (cIMLE) objective~\cite{imle} achieves exactly this: it reformulates action generation as a \emph{single-step} mapping that transforms latent noise directly into action chunks, eliminating the iterative sampling cost while retaining the expressivity of the underlying distribution. Given a training set $\{(o_t^{(i)}, A_{\mathrm{gt}}^{(i)})\}_{i=1}^{n}$, for each data point $i$ we draw $m$ (the \emph{sample factor}) independent noise vectors $\{z_{i,1},\dots,z_{i,m}\}$ and generate a set of $m$ candidate action chunks $\{\hat{A}_{i,j}\}_{j=1}^{m}$, conditioned on the corresponding VLM embedding $f_{\mathrm{VLM}}(o_t^{(i)})$:
\begin{equation}
\label{eq:sample-generation}
\hat{A}_{i,j} = G_{\theta}\!\left(f_{\mathrm{VLM}}(o_t^{(i)}),\; z_{i,j}\right), \quad j = 1,\dots,m.
\end{equation}
The cIMLE training procedure proceeds in two alternating steps. First, we perform a gradient-free \textbf{assignment step} where, for each ground-truth action $A_{\mathrm{gt}}^{(i)}$, we identify its nearest neighbour among its $m$ candidates generated from the same observation $o_t^{(i)}$ (see Eq.~\ref{eq:sample-generation}):
\begin{equation}\label{eq:assignment}
j^{*}(i) = \arg\min_{j \in \{1,\dots,m\}} \left\lVert \hat{A}_{i,j} - A_{\mathrm{gt}}^{(i)} \right\rVert_2^2.
\end{equation}
Next, in the \textbf{update step}, we minimize the distance between each ground-truth action and its assigned nearest neighbour:
\begin{equation}\label{eq:cimle}
\mathcal{L}_{\mathrm{cIMLE}}
= \frac{1}{n}\sum_{i=1}^{n}
  \left\lVert \hat{A}_{i,j^{*}(i)} - A_{\mathrm{gt}}^{(i)} \right\rVert_2^2.
\end{equation}

To see why this objective preserves multimodality, consider two illustrative regimes for the sample factor $m$: one with $m{=}1$, and another with $m>1$. Setting $m=1$ makes the nearest-neighbour selection $j^*(i)$ trivial, reducing Eq.~\ref{eq:cimle} to plain L2 regression. However, it is well-known that this objective is optimized by regression to the mean:
\begin{equation}\label{eq:regression-mean}
\mathcal{L}_{m=1}
= \mathbb{E}_{(o_t,\, A_{\mathrm{gt}})}\,
  \mathbb{E}_{z}
  \bigl\lVert G_{\theta}(f_{\mathrm{VLM}}(o_t),\, z) - A_{\mathrm{gt}} \bigr\rVert_2^2,
\end{equation}
whose minimizer ignores the noise and outputs the conditional mean:
\begin{equation}\label{eq:conditional-mean}
G_{\theta^{*}}(f_{\mathrm{VLM}}(o_t),\, z)
= \mathbb{E}\!\left[A_{\mathrm{gt}} \mid o_t\right]
\quad \text{for all } z.
\end{equation}

In effect, even when there are multiple correct actions for a given observation, this model is still optimized by averaging the multiple correct actions into one. Yet this average can lie between the modes, typically matching none of the demonstrated behaviors, making it an invalid action.

The assignment step in Eq.~\ref{eq:assignment} is what changes this. With $m>1$ candidates, each ground-truth action is matched to only its nearest candidate, allowing the other candidates to cover other modes without penalty by the cIMLE objective. Together, the outer sum ensures every mode is covered, while the nearest-neighbour assignment lets the generator map different latent codes $z$ to different modes rather than collapsing them into a single averaged action.

\begin{algorithm}[htbp]
\caption{cIMLE Training}\label{alg:cimle}
\begin{algorithmic}\raggedright
\STATE \textbf{Input:} Training batch $\{(o_t^{(i)}, A_{\mathrm{gt}}^{(i)})\}_{i=1}^{B}$, number of samples $m$, frozen VLM $f_{\mathrm{VLM}}$, action head $G_\theta$
\STATE \textbf{Output:} Updated parameters $\theta$
\STATE
\STATE // \textit{Preprocessing: Compute VLM embeddings once per batch}
\STATE Compute $f_{\mathrm{VLM}}(o_t^{(i)})$ once for all $i \in [B]$
\STATE // \textbf{Step 1: Efficient Nearest Neighbour Search (Gradient-free)}
\FOR{$i = 1$ \TO $B$}
  \STATE Sample $m$ noise vectors $\{z_{i,j}\}_{j=1}^{m} \sim \mathcal{N}(0, I)$
  \STATE Generate $m$ candidates $\hat{A}_{i,j} = G_\theta(f_{\mathrm{VLM}}(o_t^{(i)}), z_{i,j})$, $j \in [m]$
  \STATE $j^{*}(i) \leftarrow \arg\min_{j \in [m]} \lVert \hat{A}_{i,j} - A_{\mathrm{gt}}^{(i)} \rVert_2^2$ 
\ENDFOR
\STATE
\STATE // \textbf{Step 2: Training via Assigned Nearest Neighbours}
\STATE $\hat{A}_{i}^{*} \leftarrow G_\theta(f_{\mathrm{VLM}}(o_t^{(i)}),\; z_{i,\,j^{*}(i)})$ for all $i \in [B]$ \hfill $\triangleright$ \textit{With gradients}
\STATE $\mathcal{L} \leftarrow \frac{1}{B}\sum_{i=1}^{B} \lVert \hat{A}_{i}^{*} - A_{\mathrm{gt}}^{(i)} \rVert_2^2$
\STATE $\theta \leftarrow \theta - \eta\,\nabla_\theta \mathcal{L}$ \hfill $\triangleright$ \textit{Update action head only}
\end{algorithmic}
\end{algorithm}

As detailed in Algorithm~\ref{alg:cimle}, this procedure adds negligible training overhead. While the training procedure is presented sequentially for conceptual clarity, the assignment step (Step 1) is implemented in parallel across both the batch and the $m$ samples, making it highly efficient. In practice, we find that $m{=}2$ strikes an effective balance between expressivity and computational efficiency; we ablate this choice in Sec.~\ref{sec:method_analysis}. The VLM embedding $f_{\mathrm{VLM}}(o_t^{(i)})$ is computed once per sample and shared across all $m$ candidates, avoiding redundant backbone passes. Furthermore, the selection of $j^{*}(i)$ is performed without gradient tracking; only the winning candidate is recomputed for the backward pass. Consequently, cIMLE acts as a lightweight objective that transforms a multi-step VLA into a highly efficient single-step generator while explicitly optimizing for mode coverage, yielding substantial gains in inference frequency without sacrificing the policy's capacity to handle complex, multimodal tasks.

\section{Experiments}\label{sec:experiments}

In this section, we evaluate the effectiveness of IMLE-VLA through a comprehensive set of evaluations in simulation and on a real-world robot. Our goal is to demonstrate that single-step action generation via cIMLE can achieve significant gains in inference frequency while maintaining or even improving the success rates of VLA policies.

All experiments use a standard receding-horizon loop: the policy observes $o_t$, runs one forward pass to predict an action chunk (Sec.~\ref{sec:arch}), and the robot executes the first $H$ actions of that chunk \emph{open-loop} before observing again and repeating. Here $H$ is the \emph{execution horizon}, a fixed number of actions executed per forward pass, so a larger $H$ replans less often. We report three metrics: \emph{task success rate}, our primary measure of policy quality; \emph{action throughput}, the number of executable actions delivered to the robot per second, equal to inference frequency $\times\,H$; and \emph{VLA-only wall-clock} (VLA-WC), the total time spent exclusively on VLA forward passes per successful episode, excluding robot execution and other non-inference overhead. 

We benchmark IMLE-VLA against the $\pi_{0.5}$ flow-matching baseline across four primary axes:

\begin{enumerate}[label=(\roman*)]
\item Inference frequency, where we characterize raw inference frequency and action throughput; 
\item Simulated performance on the LIBERO benchmark, comprising 40 diverse manipulation tasks; 
\item Robustness under test-time distribution shift, evaluated on LIBERO-plus;
\item Real-world experiments on four tasks assessing single-step manipulation, multi-step reasoning and reactivity;
\end{enumerate}

\subsection{Inference frequency}
\label{sec:inference_frequency}

Deployment on physical robots imposes strict constraints on \textit{inference frequency}: the number of VLA forward passes completed per second, directly governing the replanning rate. Higher inference frequency enables tighter control loops, smoother motion and faster response to new visual stimuli.

We benchmark efficiency on a single NVIDIA L40S GPU using natural task-specific language prompts and the standard two-view image inputs. Our protocol reflects deployment-realistic conditions: unlike prior benchmarks using empty prompts or static buffers~\cite{ma2025realtime}, we include the mandatory cost of tokenizing the language instruction and proprioceptive state at every forward call.

Table~\ref{tab:inference_frequency} compares IMLE-VLA against several optimized implementations of the $\pi_{0.5}$~\cite{black2025pi05} VLA. The canonical JAX implementation runs at $15$\,Hz. Porting to PyTorch with \texttt{torch.\allowbreak compile}~\cite{cadene2026lerobot} and dedicated Triton kernels~\cite{ma2025realtime} yield modest gains, reaching $20$ and $25$\,Hz respectively. This is because the bottleneck is algorithmic: the $10$-step flow-matching loop imposes a fundamental ceiling that system-level optimizations alone cannot overcome. High latency is particularly impactful in physical deployments, where slow inference cycles manifest as noticeable ``stop-and-go'' pauses and reduced reactivity to environmental changes (see Sec.~\ref{sec:real_world} for concrete examples).

IMLE-VLA sidesteps this bottleneck by replacing the iterative action head with a single-step conditional generator while maintaining the exact same VLM backbone architecture. This yields an inference frequency of $55$\,Hz, a $3.67\times$ increase over the canonical JAX baseline, in turn enabling smoother robotic motion and more responsive control without reducing model capacity or backbone scale.

\begin{table}[htbp]
    \centering
    \footnotesize
    \setlength{\tabcolsep}{2pt}
    \vspace{0.9em}
    \begin{tabular}{l r r}
        \hline
        \textbf{System} & \textbf{Inf.\ Freq.\ (Hz)} $\uparrow$ & \textbf{Inf.\ Freq.\ ($\times$)} $\uparrow$ \\
        \hline
        $\pi_{0.5}$ (original JAX) & $15$ & $1.00\times$ \\
        PyTorch + compile & $20$ & $1.33\times$ \\
        Triton & $25$ & $1.67\times$ \\
        \textbf{IMLE-VLA} (Ours) & \textbf{55} & \textbf{3.67$\times$} \\
        \hline
    \end{tabular}

    \vspace{0.3em}
    \caption{\textbf{Inference frequency} measured on an NVIDIA L40S, the number of VLA forward passes per second, averaged over $50$ episodes across $10$ LIBERO-Long tasks. IMLE-VLA achieves $55$\,Hz, a $3.67\times$ increase.}
    \label{tab:inference_frequency}
    \vspace{1em}
\end{table}

\subsection{Simulation: LIBERO Benchmark}

\label{sec:libero_benchmark}

We next characterize the trade-off between inference frequency and task success on the LIBERO benchmark~\cite{liu2023libero}, which comprises 40 tabletop manipulation tasks across four suites (Spatial, Object, Goal, and Long) with a Franka Emika Panda arm. Each suite contains 10 tasks; every task is evaluated over 50 episodes following the standard protocol.

\textbf{Success rate comparison.}
Table~\ref{tab:libero_success_horizon} compares IMLE-VLA against iterative action heads~\cite{black2024pi0,black2025pi05} and speed-up strategies surveyed in Sec.~\ref{sec:related}. Inf.\ ($\times$) denotes the inference-frequency ratio relative to $\pi_{0.5}$ on matched hardware. We measure $\pi_{0.5}$, IMLE-VLA, and OpenVLA-OFT ourselves on an L40S; ratios for the remaining baselines are taken from Jeon et~al.~\cite{jeon2025shallow}, who measure on H100.

With $H{=}10$, IMLE-VLA achieves \textbf{98.0\%} average success, the highest among all compared methods, while also leading in inference frequency at $3.67\times$ $\pi_{0.5}$. The fastest competing accelerator, Shallow-$\pi_{0.5}$-L6 ($2.3\times$), is still slower than IMLE-VLA and drops to 94.5\% due to its reduced model capacity~\cite{jeon2025shallow}. OpenVLA-OFT reaches 94.5\% on the same unfiltered dataset and runs below $\pi_{0.5}$ in inference frequency ($0.69\times$), reflecting the cost of its 7B-parameter backbone. IMLE-VLA is the only method that simultaneously leads in both success rate and inference frequency.

\begin{table*}[t]
\centering

\vspace{0.9em}

\small
\setlength{\tabcolsep}{4pt}
\renewcommand{\arraystretch}{1.05}
\begin{tabular}{lrcccccc}

\toprule
\textbf{Model} & \textbf{$H$}
& \textbf{Spatial} & \textbf{Object} & \textbf{Goal} & \textbf{Long}
& \textbf{Avg} $\uparrow$
& \textbf{Inf.\ ($\times$)} $\uparrow$ \\
\midrule

$\pi_{0.5}$ \cite{black2025pi05} & 10 & 97.2 & \second{99.0} & \second{97.8} & \best{96.0} & \second{97.5} & 1.0 \\
$\pi_{0.5}$ \cite{black2025pi05} & 30 & 95.0 & 98.0 & 96.2 & 95.2 & 96.1 & 1.0 \\
$\pi_{0}$ \cite{black2024pi0} & -- & 96.8 & 98.8 & 95.8 & 85.2 & 94.2 & 1.12 \\

\midrule

OpenVLA-OFT \cite{kim2025openvlaoft} & -- & 95.2 & 94.2 & 95.2 & 93.2 & 94.5 & 0.69 \\

\midrule

CogVLA \cite{li2025cogvla} & -- & \best{99.0} & \second{99.0} & 97.0 & 95.0 & \second{97.5} & 0.8 \\
SmolVLA \cite{shukor2025smolvla} & -- & 90.0 & 96.0 & 92.0 & 71.0 & 87.3 & 1.0 \\

\midrule

Shallow-$\pi_{0.5}$-L9 \cite{jeon2025shallow} & -- & \best{99.0} & 98.0 & 97.0 & 93.0 & 96.8 & 1.7 \\
Shallow-$\pi_{0.5}$-L6 \cite{jeon2025shallow} & -- & \second{98.0} & 96.0 & 94.0 & 90.0 & 94.5 & \second{2.3} \\

\midrule

IMLE-VLA & 10 & \second{98.0} & \best{99.8} & \best{98.2} & \best{96.0} & \best{98.0} & \best{3.67} \\
IMLE-VLA & 30 & 97.2 & \second{99.0} & 96.2 & \second{95.8} & 97.1 & \best{3.67} \\

\bottomrule
\end{tabular}
\vspace{0.0em}
\caption{\textbf{LIBERO benchmark.} Success rates (SR, \%) on four LIBERO suites (50 episodes per task) and their average. $H$ denotes the execution horizon (actions executed open-loop per forward pass; ``--'' = not specified). Inf.\ ($\times$) is the inference frequency ratio relative to $\pi_{0.5}$. Best in \textbf{bold}, second-best \underline{underlined}. IMLE-VLA achieves the highest average success rate ($98.0\%$ at $H{=}10$) while simultaneously leading in inference frequency ($3.67\times$).}
\label{tab:libero_success_horizon}
\end{table*}

\begin{figure}[H]
    \centering
    \includegraphics[width=0.7\linewidth]{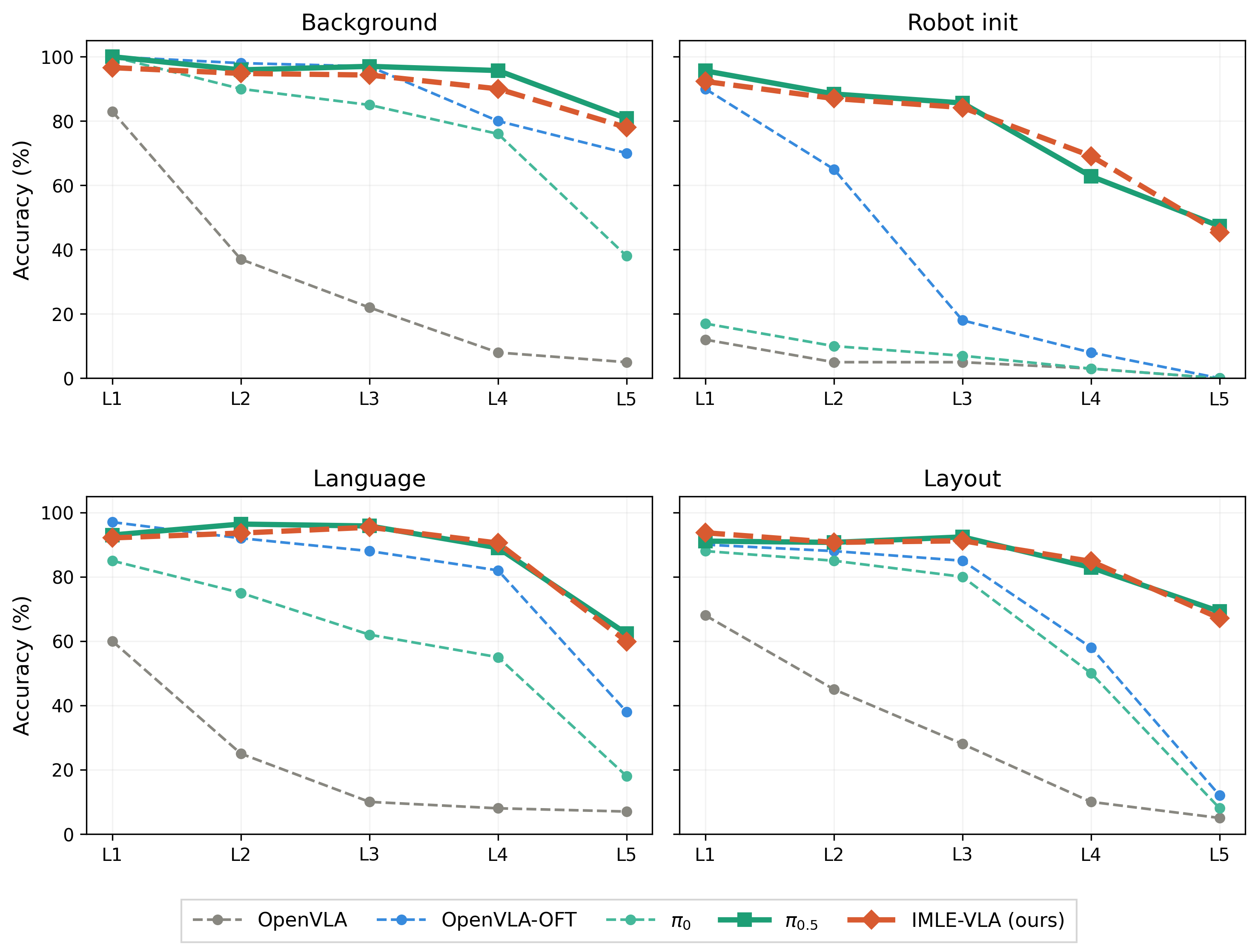}
    \caption{\textbf{LIBERO-plus robustness under distribution shift.} Success across four perturbation dimensions, each at five difficulty levels (L1--L5). Unlike baselines, which degrade sharply as severity increases, IMLE-VLA matches $\pi_{0.5}$'s robustness throughout.}
    \label{fig:libero_plus}
  \end{figure}

\textbf{Robustness under distribution shift.} To verify that our gains generalize beyond LIBERO's training distribution, we additionally evaluate on \textbf{LIBERO-plus}~\cite{fei25libero-plus}, which applies systematic test-time perturbations to the LIBERO tasks across four axes (background, robot initial state, language, layout) at five severity levels (L1--L5). Across all axes and severities, IMLE-VLA retains $\pi_{0.5}$'s robustness, while all other baselines exhibit larger drops in success rate as severity increases (Fig.~\ref{fig:libero_plus}). Consistent with the L1-regression limitation discussed in Sec.~\ref{sec:related}~\cite{kim2025openvlaoft}, OpenVLA-OFT's success rate degrades sharply. IMLE-VLA's cIMLE objective instead enforces mode coverage \emph{by construction}~\cite{imle}, avoiding this failure mode.

\textbf{Action throughput.}
Table~\ref{tab:action_throughput_h10} reports action throughput (inference frequency $\times\,H$) across configurations. At $H{=}10$, IMLE-VLA already delivers $3.67\times$ more actions per second than the $\pi_{0.5}$ baseline, purely from replacing the flow-matching head. At $H{=}30$, the single-step inference advantage compounds with the longer horizon, yielding $11.0\times$ higher throughput.

\textbf{VLA-only wall-clock (VLA-WC).} Table~\ref{tab:policy_time} reports VLA-WC across horizons, averaged over successful LIBERO episodes. At $H{=}10$ the number of forward passes per episode is nearly identical to $\pi_{0.5}$, yet VLA-WC drops by $3.6\times$ due to the lighter single-step action head. At $H{=}30$ the forward-pass count itself falls by ${\sim}3\times$; combined with the faster per-pass inference, this yields an overall $10.5\times$ reduction.

\begin{table}[H]
    \centering
    \footnotesize
    \setlength{\tabcolsep}{3.4pt}
    \vspace{0.0em}
    \begin{tabular}{l r r r}
        \hline
        \textbf{System} & \textbf{$H$} & \textbf{Action Throughput (Hz)} & \textbf{($\times$)} \\
        \hline
        $\pi_{0.5}$ (JAX) & 10 & 150 & 1.00$\times$ \\
        PyTorch + \texttt{torch.compile} & 10 & 200 & 1.33$\times$ \\
        Triton & 10 & 250 & 1.67$\times$ \\
        Ours & 10 & 550 & 3.67$\times$ \\
        Ours & 30 & 1{,}650 & 11.00$\times$ \\
        \hline
    \end{tabular}
    \vspace{0.3em}
    \caption{\textbf{Action throughput} in LIBERO simulation, defined as inference frequency $\times$ execution horizon $H$. ($\times$) denotes the throughput ratio relative to $\pi_{0.5}$ at $H{=}10$. At $H{=}10$, IMLE-VLA delivers $3.67\times$ higher throughput purely from single-step generation; at $H{=}30$ this compounds to $11.0\times$.}
    \label{tab:action_throughput_h10}
\end{table}

\begin{table}[H]
    \centering
    \vspace{0.9em}

    \footnotesize
    \setlength{\tabcolsep}{3pt}
    \begin{tabular}{l r r r}
        \hline
        \textbf{System} & \textbf{Fwd.\ Passes} & \textbf{VLA-WC (s)} & \textbf{VLA-WC ($\times$)} \\
        \hline
        $\pi_{0.5}$ (JAX, $H{=}10$) & 15.4 & 1.03 & 1.0$\times$ \\
        Ours ($H{=}10$) & 15.5 & 0.28 & 3.6$\times$ \\
        Ours ($H{=}30$) & 5.4 & 0.10 & 10.5$\times$ \\
        \hline
    \end{tabular}
    \vspace{0.3em}
    \caption{\textbf{VLA-only wall-clock (VLA-WC).} Wall-clock on VLA forward passes per successful episode (see Sec.~\ref{sec:experiments}). \emph{Fwd.\ Passes}: average forward passes per episode. At $H{=}10$, IMLE-VLA reduces VLA-WC by $3.6\times$; at $H{=}30$ fewer passes compound with faster inference for $10.5\times$ overall.}
    \label{tab:policy_time}
\end{table}

\subsection{Method Analysis}
\label{sec:method_analysis}

\begin{figure}[H]
    \centering
    \includegraphics[width=0.5\linewidth]{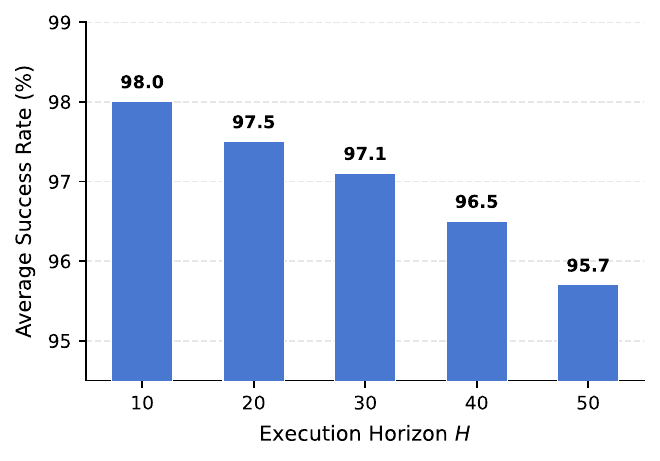}
    \caption{\textbf{LIBERO horizon ablation.} IMLE-VLA success averaged across all four LIBERO suites at varying execution horizons. $H{=}10$ achieves 98.0\%; larger horizons reduce reactivity and degrade success, while $H{=}30$ offers the best throughput--accuracy balance.}
    \label{fig:horizon_ablation}
\end{figure}

\textbf{Horizon ablation.} Each replan cycle consists of one forward pass followed by $H$ open-loop actions. Because IMLE-VLA's forward pass is $3.67\times$ faster, the saved inference time can be reinvested into a longer horizon without exceeding $\pi_{0.5}$'s cycle time, making larger horizons practical. However, extending $H$ still trades \emph{reactivity} for \emph{throughput}: the robot executes more actions before conditioning on a new observation. Figure~\ref{fig:horizon_ablation} reports IMLE-VLA's average LIBERO success across all four suites at varying horizons. At $H{=}10$ the policy achieves 98.0\% average success, and performance degrades gradually as $H$ grows. $H{=}30$ offers the best throughput--accuracy balance, still reaching 97.1\% while delivering $11.0\times$ higher action throughput (Table~\ref{tab:action_throughput_h10}), and outperforming $\pi_{0.5}$ at the same horizon (96.1\%).

\textbf{cIMLE sample factor ablation.} The sample factor $m$ controls how many candidate action chunks are generated per training example during the cIMLE assignment step (Eq.~\ref{eq:sample-generation}). When $m{=}1$ only one candidate exists, so the assignment is trivial and cIMLE reduces to standard L2 regression with no mode-covering guarantee.

\begin{figure}[htbp]
    \centering
    \includegraphics[width=0.4\linewidth]{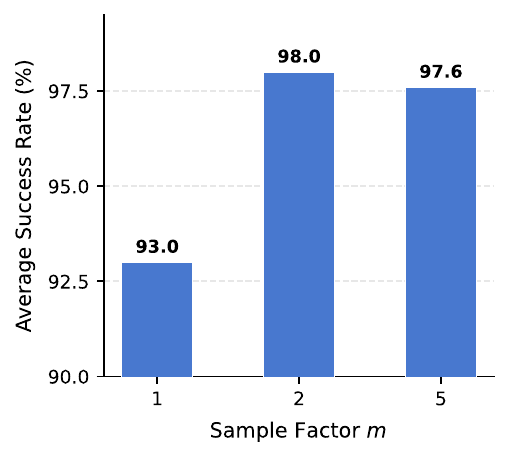}
    \caption{\textbf{cIMLE sample factor ablation.} LIBERO success averaged across all four suites at $H{=}10$ for varying sample factor $m$. $m{=}1$ reduces cIMLE to standard regression (no mode-covering) and drops sharply; $m{=}2$ and $m{=}5$ are nearly identical, so we use $m{=}2$.}
    \label{fig:sample_factor_ablation}
\end{figure}

Figure~\ref{fig:sample_factor_ablation} reports LIBERO success across all four suites at $H{=}10$ for $m \in \{1,2,5\}$. Setting $m{=}1$ produces a clear performance drop, confirming that a pure regression objective is insufficient for modeling complex action distributions, an observation consistent with the limitation acknowledged by OpenVLA-OFT~\cite{kim2025openvlaoft} (see Sec.~\ref{sec:related}). With $m{=}2$ the mode-covering property activates and performance recovers substantially; increasing to $m{=}5$ yields nearly identical results, indicating diminishing returns. We therefore use $m{=}2$ throughout, as it achieves strong performance while minimizing the per-step training cost of generating additional candidates.

\subsection{Real-world experiments}\label{sec:real_world}

\begin{figure}[tbp]
    \centering
    \vspace{0.3em}

    \includegraphics[width=0.6\columnwidth]{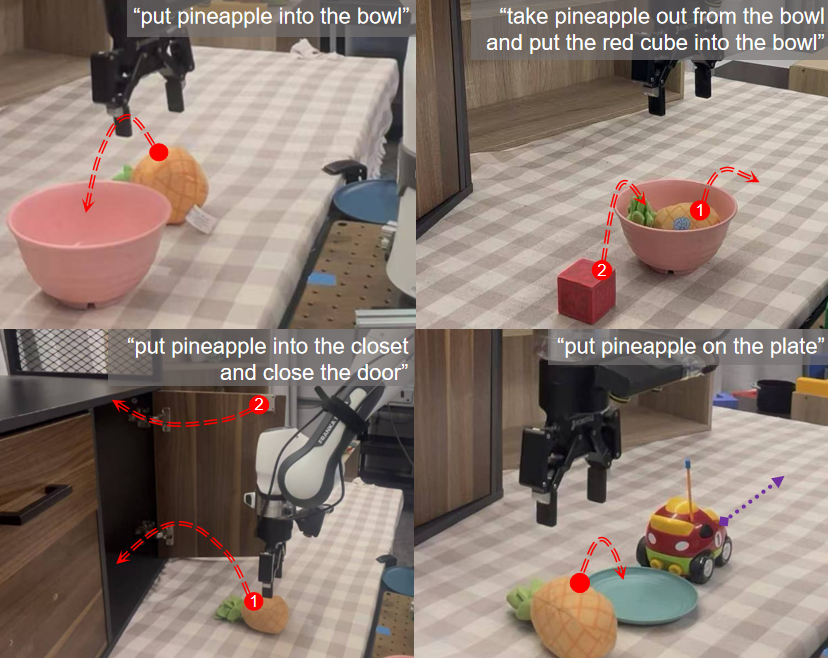}

    \vspace{0.3em}
    \caption{\textbf{Real-world experiments.} We evaluate on three categories: \textbf{Single-Step Manipulation}, \textbf{Multi-Step Sequential Reasoning} and \textbf{Dynamic Reactivity}. IMLE-VLA generates faster, smoother, and more successful actions across all three categories.}
    \label{fig:realworld_tasks}
    \end{figure}

We evaluate IMLE-VLA against the $\pi_{0.5}$ baseline (canonical JAX implementation) on a Franka Emika Panda robot with an NVIDIA A6000 GPU, using two camera views (wrist and scene). Both policies are trained on DROID~\cite{khazatsky2024droid} and deployed zero-shot on four tabletop manipulation tasks (Figure~\ref{fig:realworld_tasks}), with success (task completion) reported over 20 episodes per task. $\pi_{0.5}$ runs at 15\,Hz control frequency, while IMLE-VLA operates at 55\,Hz. Unlike LIBERO's deterministic contact dynamics, real-world manipulation introduces physical uncertainty (e.g.\ grasp slippage, surface friction variation), requiring more frequent replanning. We adopt the $\pi_{0.5}$ default horizon ($H{=}8$) and extend IMLE-VLA to $H{=}12$, since the faster inference compensates for the longer horizon while still yielding a shorter wall-clock replan interval than $\pi_{0.5}$. IMLE-VLA outperforms $\pi_{0.5}$ in every task while reducing the VLA-only wall-clock by $3.9\times$--$6.6\times$ (Table~\ref{tab:realrobot_static}) and completing episodes approximately $2\times$ faster end-to-end (supplementary videos). The four tasks span single-step manipulation, multi-stage sequential reasoning, and reactivity to a dynamic environment.

\begin{itemize}
    \item \textbf{Pineapple in bowl} (single-step): the robot grasps a squishy pineapple toy from the table and places it into a bowl.
    \item \textbf{Swap pineapple and cube} (multi-step): a bowl contains a squishy pineapple with a red cube beside it. The robot must remove the pineapple from the bowl, then place the red cube inside.
    \item \textbf{Put pineapple in cabinet} (multi-step): a cabinet door is already open with a squishy pineapple in front of it. The robot must pick up the pineapple, place it inside the cabinet, and close the door.
    \item \textbf{Pineapple on moving plate} (reactive): a remote-controlled car drives an attached plate across the table; the robot must place the pineapple on the moving plate, requiring continuous target adjustment.
\end{itemize}
    
\begin{table}[t]
    \centering
    \footnotesize
    \setlength{\tabcolsep}{2.8pt}
    \vspace{0.9em}
    \begin{tabular}{@{}l cc cc c@{}}
        \toprule
        & \multicolumn{2}{c}{\textbf{Success} $\uparrow$} & \multicolumn{2}{c}{\textbf{VLA-WC (s)} $\downarrow$} & \\
        \cmidrule(lr){2-3} \cmidrule(lr){4-5}
        \textbf{Task} & \textbf{Ours} & \textbf{$\pi_{0.5}$} & \textbf{Ours} & \textbf{$\pi_{0.5}$} & \textbf{Jerk ($\times$)} \\
        \midrule
        Pineapple in bowl         & 19 & 15 & 18.56 & 123 & $2.7\times$ \\
        Swap pineapple \& cube    & 18  & 15 & 31.1  & 178.3  & $2.2\times$ \\
        Pineapple in cabinet      & 15  & 12 & 99.4  & 389.4  & $3.0\times$ \\
        Pineapple on moving plate & 16  & 12 & 16.7  & 89.5   & $2.7\times$ \\
        \bottomrule
    \end{tabular}
    \vspace{0.3em}
    \caption{\textbf{Real-world experiment results} on a Franka Emika Panda with an NVIDIA A6000 (20 episodes per task). Success: successful episodes out of 20. VLA-WC: wall-clock on VLA forward passes per successful episode, averaged over successful episodes per task. Jerk ($\times$): ratio of $\pi_{0.5}$'s mean proprioceptive jerk to ours (higher means smoother motion relative to the baseline; see Sec.~\ref{sec:real_world}). IMLE-VLA outperforms $\pi_{0.5}$ on every task, reduces VLA-WC by $3.9\times$--$6.6\times$, and produces $2.2\times$--$3.0\times$ lower jerk.} 
    \label{tab:realrobot_static}
\end{table}

\textbf{Motion quality.} IMLE-VLA produces markedly smoother motion than $\pi_{0.5}$: its faster inference minimizes the time the robot idles while the model computes, and its longer execution horizon means fewer replans, together eliminating the visible stop-wait-execute pauses and abrupt directional corrections that $\pi_{0.5}$ exhibits at each replan. To quantify this, we measure \emph{proprioceptive jerk}, the third backward finite difference of the measured joint positions (7 arm joints, gripper excluded) normalized by $dt^{3}$, averaged over joints and steps; IMLE-VLA reduces mean jerk by $2.2\times$--$3.0\times$ across all four tasks (Table~\ref{tab:realrobot_static}). The difference is apparent in the supplementary videos. On \emph{swap pineapple and cube}, $\pi_{0.5}$ changes direction repeatedly before each placement, while IMLE-VLA follows a direct trajectory and finishes faster. The shorter replan interval also enables recovery when something goes wrong: on \emph{pineapple in cabinet}, when the pineapple slips from IMLE-VLA's gripper during insertion, the robot quickly detects the failure and re-grasps. Meanwhile, $\pi_{0.5}$'s slower replanning introduces frequent abrupt corrections that significantly slow task completion.

\textbf{Reactive task.} The dynamic task (\emph{pineapple on moving plate}) evaluates the ability to react to a continuously changing environment. In this setting, IMLE-VLA significantly outperforms $\pi_{0.5}$. Although IMLE-VLA executes a longer action horizon, its faster inference leads to a much shorter wall-clock replanning cycle. As a result, the policy effectively closes the perception--action loop more frequently in real time, allowing the robot to continuously track and adapt to the plate's motion.

This distinction is particularly important in dynamic scenes, where delays between perception and action quickly degrade performance. By minimizing idle time between replans, IMLE-VLA reduces the latency between observing the plate and responding to its updated position, enabling smoother interception and grasp attempts. In contrast, $\pi_{0.5}$ replans more slowly and therefore acts on increasingly stale observations. By the time a new action sequence is generated, the plate has already moved, causing the robot to repeatedly reach toward outdated target locations and ultimately fail to complete the task.

\section{Conclusion}
\label{sec:conclusion}

We presented \textbf{IMLE-VLA}, a framework that replaces the iterative flow-matching action head of VLAs with a single-step conditional generator trained via the cIMLE objective. By eliminating multi-step sampling while explicitly optimizing for mode coverage, IMLE-VLA achieves a $3.67\times$ increase in inference frequency and up to $11.0\times$ higher action throughput over the $\pi_{0.5}$ baseline. On the LIBERO benchmark, IMLE-VLA attains the highest average success rate (98.0\%) among all compared methods while simultaneously leading in inference frequency. In real-world experiments on a Franka Emika Panda, the higher inference frequency produces smoother motion and reduces VLA-only wall-clock by $3.9\times$--$6.6\times$, while outperforming $\pi_{0.5}$ on every task. Because our approach modifies only the action head, it serves as a drop-in replacement for the iterative heads in existing VLAs, a promising direction toward even faster generalist robot policies.

\section*{ACKNOWLEDGMENT}

This research was enabled in part by NSERC, the Canada Foundation for Innovation, the Canada CIFAR AI Chairs program, the BC DRI Group and the Digital Research Alliance of Canada. The first author dedicates this work to the memory of his beloved grandfather.

\bibliographystyle{IEEEtran}
\bibliography{references}

\end{document}